%% file: main.tex
\documentclass[runningheads]{llncs}
\usepackage{graphicx}
\usepackage[ruled,linesnumbered]{algorithm2e}
  \SetCommentSty{commentstyle}
\usepackage[style=base]{caption}
\usepackage[labelformat=simple]{subcaption}
	
\usepackage{booktabs}
\usepackage{multirow}
\usepackage{tikz}
\usetikzlibrary{positioning,shapes,calc,patterns}
\tikzset{
	t/.style={->},
	cross/.style={cross out,draw,minimum size=2*(#1-\pgflinewidth),inner sep=0pt,outer sep=0pt,thick},
	cross/.default={1mm},
	plus/.style={cross,rotate=45,thin},
	p/.style={plus},
	p1/.style={rectangle,fill=niceorange,minimum size=1mm,inner sep=0mm},
	p2/.style={circle,fill=niceblue,minimum size=1mm,inner sep=0mm},
	c/.style={cross},
}
\usepackage{pgfplots}
\pgfplotsset{compat = 1.8}
\colorlet{ycolor}{green!40!blue}
\colorlet{xcolor}{red}
\usepackage{amsmath,amssymb}
\usepackage{csvsimple}
\usepackage{xcolor}
\usepackage{hyperref}

\newcommand{\NN}{\ensuremath{\mathcal{N}}\xspace}

\newcommand{\M}{\ensuremath{\mathcal{M}}\xspace}
\newcommand{\Vals}[1][]{\ensuremath{\mathcal{V}_{#1}}\xspace}
\newcommand{\vc}{\ensuremath{\vec{c}}\xspace}
\newcommand{\vr}{\ensuremath{\vec{r}}\xspace}
\newcommand{\vp}{\ensuremath{\vec{p}}\xspace}
\newcommand{\vx}{\ensuremath{\vec{x}}\xspace}

\newcommand{\vs}{\ensuremath{\vec{s}}\xspace}
\newcommand{\onebox}{\ensuremath{B}\xspace}
\newcommand{\boxes}{\ensuremath{\mathcal{B}}\xspace}
\newcommand{\statmon}{\ensuremath{s_\mathit{monitor}}\xspace}
\newcommand{\statnet}{\ensuremath{s_\mathit{network}}\xspace}
\newcommand{\statsample}{\ensuremath{s_\mathit{samples}}\xspace}
\newcommand{\statmonStar}{\ensuremath{\statmon^*}\xspace}
\newcommand{\statnetStar}{\ensuremath{\statnet^*}\xspace}
\newcommand{\statsampleStar}{\ensuremath{\statsample^*}\xspace}
\newcommand{\cc}[1]{\multicolumn{1}{c @{\hspace*{4mm}}}{#1}}
\newcommand{\dbox}{d_{\boldsymbol+}}

\newcommand{\algref}[1]{Algorithm~\ref{algo:#1}\xspace}
\newcommand{\figref}[1]{Fig.~\ref{fig:#1}\xspace}
\newcommand{\tabref}[1]{Table~\ref{tbl:#1}\xspace}

\definecolor{niceblue}{rgb}{0.0, 0.45, 0.7}
\definecolor{bluegreen}{rgb}{0.0, 0.6, 0.5}
\definecolor{niceorange}{rgb}{0.9, 0.6, 0.0}
\definecolor{vermilion}{rgb}{0.8, 0.4, 0.0}
\definecolor{redpurple}{rgb}{0.8, 0.6, 0.7}
\definecolor{niceyellow}{rgb}{0.95, 0.9, 0.25}
\definecolor{skyblue}{rgb}{0.35, 0.7, 0.9}
\definecolor{nicegray}{rgb}{0.204, 0.282, 0.275}

\newcommand{\classes}{\ensuremath{\mathcal{Y}}\xspace}
\newcommand{\class}{\ensuremath{y}\xspace}
\newcommand{\traindata}{\ensuremath{\mathcal{X}}\xspace}
\newcommand{\observe}{\ensuremath{\mathop{\mathbf{observe}}}}
\newcommand{\get}{\ensuremath{\mathop{\mathbf{get}}}}
\newcommand{\classify}{\ensuremath{\mathop{\mathbf{classify}}}}
\newcommand{\evaluate}{\ensuremath{\mathop{\mathbf{evaluate}}}}
\newcommand{\adaptMonitor}{\ensuremath{\mathop{\mathbf{adaptMonitor}}}}
\newcommand{\adaptModel}{\ensuremath{\mathop{\mathbf{adaptModel}}}}
\newcommand{\askAuthority}{\ensuremath{\mathop{\mathbf{askAuthority}}}}
\newcommand{\collect}{\ensuremath{\mathop{\mathbf{collect}}}}
\newcommand{\build}{\ensuremath{\mathop{\mathbf{buildMonitor}}}}
\newcommand{\monitor}{\ensuremath{\mathop{\mathbf{monitor}}}}
\newcommand{\asgn}{\ensuremath{\gets}\xspace}
\newcommand{\transpose}{^\mathrm{T}\xspace}
\newcommand{\cI}{{\color{niceblue}\ensuremath{\bullet}}\xspace}
\newcommand{\cII}[1][\raisebox{0.3mm}]{{\color{niceorange}#1{\scalebox{0.5}{\ensuremath{\blacksquare}}}}\xspace}

\newcommand*\rboxed[1]{\tikz[baseline=(char.base)]{
            \node[shape=rectangle,rounded corners,draw=bluegreen,inner sep=1.5pt, text=bluegreen] (char) {#1};}}

\begin{document}
\title{Into the Unknown: Active Monitoring \\of Neural Networks\thanks{%
This is a post-peer-review, pre-copyedit version of an article published in RV 2021.
The final authenticated version is available online at: \url{https://doi.org/10.1007/978-3-030-88494-9_3}.
}}
\author{Anna Lukina%
\inst{1}\orcidID{0000-0001-9525-0333} \and
Christian Schilling%
\inst{2}\orcidID{0000-0003-3658-1065} \and
Thomas A. Henzinger\inst{1}\orcidID{0000-0002-2985-7724}}
\authorrunning{A. Lukina et al.}
\institute{Institute of Science and Technology Austria \and
University of Konstanz, Germany\\
\email{\{anna.lukina,thomas.henzinger\}@ist.ac.at}\\
\email{christian.schilling@uni-konstanz.de}}
\maketitle
\begin{abstract}
Neural-network classifiers achieve high accuracy when predicting the class of an input that they were trained to identify. Maintaining this accuracy in dynamic environments, where inputs frequently fall outside the fixed set of initially known classes, remains a challenge. The typical approach is to detect inputs from novel classes and retrain the classifier on an augmented dataset. However, not only the classifier but also the detection mechanism needs to adapt in order to distinguish between newly learned and yet unknown input classes. To address this challenge, we introduce an algorithmic framework for active monitoring of a neural network. A monitor wrapped in our framework operates in parallel with the neural network and interacts with a human user via a series of interpretable labeling queries for incremental adaptation. In addition, we propose an adaptive quantitative monitor to improve precision. An experimental evaluation on a diverse set of benchmarks with varying numbers of classes confirms the benefits of our active monitoring framework in dynamic scenarios.
\end{abstract}

\section{Introduction}

Automated classification is an essential part of numerous modern technologies and one of the most popular applications of deep neural networks \cite{Liu2017}. Neural-network image classifiers have fast-forwarded technological development in many research areas, e.g., automated object localization as a stepping stone to successful real-world robotic applications \cite{Tobin2017}.
Such applications require a high level of reliability from the neural networks.

However, when deployed in the real world, neural networks face a common problem of novel input classes appearing at prediction time, leading to possible misclassifications and system failures. For example, consider a scenario of a neural network used for labeling inputs and making decisions about the next actions for an automated system with limited human supervision: a robot assistant learning to recognize objects in a new home. Assume the neural network is trained well on a dataset containing examples of a finite set of classes. However, after this robot is deployed in the real home, novel classes of objects can appear and confuse the neural network. The inherent misclassifications can stay undetected and accumulate over time, eventually reducing overall accuracy. 

The likelihood of severe system damage increases with the frequency and diversity of novel input classes. 
Typically, this risk is addressed by detecting novel inputs, augmenting the training dataset, and retraining the classifier from scratch \cite{ParisiKPKW19}. This procedure is not only inefficient, but also leaves the system vulnerable until such a dataset has been collected.
Techniques to incrementally adapt classifiers at prediction time are beneficial for improving accuracy in real-world applications \cite{RoyerL15,RebuffiKSL17}. They, however, do not provide desired interpretability for the human. Approaches to run-time monitoring of neural networks were therefore introduced \cite{RahmanCD21}. In particular, approaches based on abstractions \cite{ChengNY19,henzinger2020outside,chen2021monitoring,wu2021customizable} proved to be effective at detecting novel input classes. In addition, they provide transparency of neural-network monitoring.

Crucially, these monitors are constructed offline and remain static at prediction time.
Functionalities they are still lacking are distinguishing between ``known'' and ``unknown'' novelties and selectively adapting at prediction time.

We propose an active monitoring framework for neural networks that detects novel input classes, obtains the correct labels from a human authority, and adapts the neural network and the monitor to the novel classes, all at prediction time.
The framework contains a mechanism for automatic switching between monitoring and adaptation based on run-time statistics.
Adaptation consists of either learning new classes (when enough data has been collected) or retraining with more up-to-date information (when the run-time performance is unsatisfactory), where retraining is applied to the network and the monitor independently.
A trained neural-network model accompanied by our framework, as an external observer and mediator between the neural network and the human, achieves improved transparency of operation through informative interaction.

Furthermore, we propose a new monitor designed for the adaptive setting.
Introducing a quantitative metric at the hidden layers of the neural network, the monitor timely warns about inputs of novel classes and reports its own confidence to the authority. This allows for assessing the need of model adaptation.
The quantitative metric allows for easy adaptation at prediction time to newly introduced labels and successfully maintains overall classification accuracy on inputs of known and previously novel classes combined. As such, our framework is an interactive and interpretable tool for informed decision making in neural-network based applications.
We summarize the contributions of this paper below.

\begin{enumerate}
    \item We propose an automatic framework with two modes, monitoring and adaptation, that operates in parallel with the original neural network and adapts the monitor to novel input classes at prediction time.
    \item We propose a quantitative metric to measure the confidence of the novelty detection and to guide the monitor refinement. In contrast to traditional qualitative monitoring, which judges whether or not an observed input-output pair of the network is reliable, our {\em quantitative} monitoring approach computes a numerical ``reliability score'' for each observed input-output pair. The score corresponds to the distance of values in feature space at training and prediction time.
    \item We provide an experimental evaluation on a diverse set of image-classification benchmarks, demonstrating the effectiveness of the framework for achieving high monitor precision over time. Given a fixed budget of times the monitor can query the authority for a label, our monitoring approach adapts to the available classes and consumes the budget more effectively.
\end{enumerate}

After reviewing the related work in the next section, we provide the background and assumptions used throughout the paper in Section~\ref{sec:background}. We describe our quantitative monitor and the process of adaptation in Section~\ref{sec:approach}. We report on our experimental results in Section~\ref{sec:experiments} and conclude in Section~\ref{sec:discussion}.

\section{Related work}

\paragraph{Novelty detection.}

Gupta and Carlone consider neural networks that estimate human poses, for which they propose a domain-specific monitoring algorithm trained on perturbed inputs \cite{GuptaC20}. Our framework is not limited to any specific domain of images. 
Common novelty-detection approaches \cite{PimentelCCT14} examine the input sample distribution \cite{KnorrN97}, which is computationally heavier at run-time than our monitors.
Several approaches monitor the neuron valuations and compare to a ``normal'' representation of those valuations per class, obtained for a training dataset:
the patterns of neuron indices with highest values \cite{SchultheissKFD17} or positive/non-positive values \cite{ChengNY19}, and a box abstraction \cite{henzinger2020outside,wu2021customizable,IbrahimN21}. These monitors are purely qualitative and hence not adaptive, in contrast to our metric-based monitor.

\paragraph{Anomaly detection.}

There exist other directions for detecting more general anomalous behavior, not necessarily only novel classes.
In \emph{selective classification}, an input is rejected based on a (quantitative) confidence score, already at training time \cite{GeifmanE17}.
The probably best-known approach classifies based on the \emph{softmax} score \cite{GuoPSW17,HendrycksG17}, which is shown to be limited in effect \cite{GalG16}.
Approaches to \emph{failure prediction} aim to identify misclassifications of \emph{known} classes \cite{ZhangWFHP14}.
\emph{Domain adaptation} techniques detect when the underlying data distribution changes, which is necessary for statistical methods to work reliably \cite{RedkoMHSB19}.
Notably, Royer and Lampert show that correlations in the data distribution can be exploited to increase a classifier's precision \cite{RoyerL15}; while that approach applies to arbitrary classifiers in an unsupervised setting, it cannot deal with unknown classes.
Sun and Lampert study the detection of \emph{out-of-spec situations}, when classes do not occur with the expected frequency \cite{SunL20}.
An important aspect of domain adaptation, \emph{transfer learning} \cite{PanY10,TanSKZYL18}, is challenging online~\cite{ZhaoH10}.

\paragraph{Continuous / incremental learning.}

A central obstacle in incremental learning is \emph{catastrophic forgetting}: the classifier's precision for known classes decreases over time \cite{McCloskeyC89}.
We mitigate that obstacle by maintaining a sample of the training data and tuning the model on demand.
Mensink et al.\ find that a simple nearest-class-mean (NCM) classifier (mapping an input to feature space and choosing the closest centroid of all known classes) is effective \cite{MensinkVPC13}; they also consider multiple centroids per class, as we do, but they use the Mahalanobis distance in contrast to our more lightweight distance.
Guerriero et al.\ extend that idea to nonlinear deep models, where the focus is on efficiency to avoid constant retraining \cite{GuerrieroCM18}; we also delay retraining (network and monitor) until precision deteriorates.
Rebuffi et al.\ extend the NCM classifier for \emph{class-incremental learning} with fixed memory requirements \cite{RebuffiKSL17}. That learning approach, working in a completely supervised scenario, retrains the neural network using sample selection/herding and rehearsal. These ideas could also be integrated in our framework, but a representative sampling for our monitor is harder to obtain.
Similar to the NCM approach is the proposal by Mandelbaum and Weinshall to obtain a confidence score using a $k$-nearest-neighbor distance based on the Euclidean distance with respect to the training dataset, for which they require to modify the training procedure \cite{MandelbaumW17}; we do not need access to the training procedure and we experimentally found that the Euclidean distance is not suitable for networks with different scales at different neurons.

\paragraph{Active learning.}

Our approach is inspired by active learning. Active learning aims to maximize prediction accuracy even on unseen data by detecting the most representative novel inputs to label and incrementally retraining the neural network on a selected sample of labeled novelties \cite{Settles2012}. In contrast, the performance of our framework is measured primarily by the run-time precision of the monitor. We therefore use the incrementally retrained neural network solely for the monitor adaptation in parallel with the original model.
Our quantitative monitor also reasons about the feature space of the neural network (and not the input space).

An essential idea behind active learning is that, when selecting the training data systematically, fewer training samples are needed; this selection is usually taken at run-time by posing labeling queries to an authority \cite{Settles2012}.
Our approach follows the spirit of \emph{selective sampling}, where data comes from a stream, from the \emph{region of uncertainty} \cite{CohnAL94}.
Das et al.\ follow a statistical approach to outlier detection adapting to the reactions of the authority~\cite{DasWDFE16}.

In an \emph{open world} setting, novel classes have to be detected on the fly and the classifier needs to be adapted accordingly.
This setting is first approached in \cite{BendaleB15} using an NCM classifier and in \cite{BendaleB16} with a softmax score.
More recently, Mancini et al.\ propose a deep architecture for learning new classes dynamically \cite{ManciniK0JC19}.
Wagstaff et al.\ argue that two main obstacles in this setting are the cold starts and the cost of having the classifier in the loop~\cite{WagstaffL20}.

\section{Background and assumptions}
\label{sec:background}

In this paper, we deal with neural networks, which we denote by \NN.
For simplicity we present the concepts assuming a single feature layer $\ell$ of the network, but they generalize to multiple feature layers in a straightforward way.
A \emph{monitor} is a function that takes both an input and the prediction of a classifier, and then assesses whether that prediction is correct.
The monitor raises a warning if it suspects that the prediction is incorrect.
The assessment can be qualitative (``yes'' or ``no'') or quantitative (expressing the confidence of the monitor).
We write $\vx$ for an unlabeled data point, \traindata for a (possibly labeled) dataset, $\class \in \classes$ for a class in a set of classes, and $(\vx, \class)$ for a labeled data point.

\paragraph{Observing feature layers:}
We are given a trained neural network \NN and a labeled dataset \traindata with classes \classes (which is not necessarily the dataset that \NN was trained on).
When we \emph{observe a feature layer $\ell$} for some input, we obtain the corresponding neuron valuations at layer $\ell$, which we regard as high-dimensional vectors.
We can thus compute the set of neuron valuations \Vals[\class] for each class~$\class \in \classes$ and each corresponding sample in \traindata.

\paragraph{Performance metrics:}
As conventionally used for assessing the performance of classifiers and monitors, we compute the \emph{precision score}.
For a classifier this is the ratio of correct classifications over all predictions, while for a monitor this is the ratio of correct warnings (true positives, $\mathtt{TP}$) over the total number of warnings (including false positives, $\mathtt{FP}$): $\mathtt{TP/(TP+FP)}$.
At run-time we can only compute the precision score based on samples that we know the ground truth for, i.e., samples reported by the monitor and subsequently labeled by an authority.

\paragraph{Hyperparameters:}
We define the \emph{model performance threshold} \statnetStar as $95$\% of the precision score of the original neural-network model \NN on a test dataset (with classes known to \NN), which we use for making decisions about model adaptation.
The parameter \statsampleStar is the number of collected and labeled data samples of a novel class sufficient for incremental adaptation of the model to this class, which we set to $\statsampleStar = 0.05\cdot|\traindata|/|\classes|$ for an initially given dataset.
The parameter \statmonStar is the desired precision threshold of the monitor at run-time, which we set to $0.9$.
The parameters $d^*(\class)$ (for each class $\class\in\classes$) are thresholds for refining the set of inputs detected by the monitor, initialized to~$1$.

\paragraph{Assumptions:}
In this work we make a number of assumptions.
First, we assume the availability of an authority that assigns the correct label for any input that is requested.
While a human can play this role in many cases, in certain applications like medical image processing such an authority does not necessarily exist.
Second, in our experimental setup we assume that the authority is available in real time.
We also occasionally adapt the monitor or retrain the neural network.
While faster than building from scratch, this takes a non-negligible amount of time.
In time-critical applications one would need to delay these interactions and adaptations accordingly.
Third, neural networks require a large amount of data points in order to learn new classes.
In our evaluation there is sufficient data available. Still, there are approaches that work with only few samples \cite{BendreTN20,LuGYZ20}.

\section{Approach}
\label{sec:approach}

We design our monitoring framework to achieve high precision in detecting novel classes without depressing the learned model's run-time performance. To address this trade-off, our framework operates in stages, switching between monitoring and adaptation. This procedure is based on parallel composition of two components: a dynamically adapted copy of the original neural network and a monitor that originally knows the same classes as the network.
During monitoring, inputs to the network that are reported by the monitor are submitted to an authority for assigning the correct label. From that, precision scores for both the monitor and the neural network are assessed for whether adaptation is required. During adaptation, depending on the assessment, the neural network or the monitor is incrementally adjusted, or they are retrained in order to learn an unknown class.

\subsection{Quantitative monitor}
\label{subsec:monitor_init}

\begin{figure}[t!]
	\centering
	\include{tikz/initialization_raw}
	\begin{subfigure}[b]{0.49\linewidth}
		\centering
		\include{tikz/initialization_pca}
		\vspace*{-5mm}
		\caption{}
		\label{fig:raw_pca}
	\end{subfigure}%
	\begin{subfigure}[b]{0.49\linewidth}
		\centering
		\include{tikz/initialization_clustered}
		\vspace*{-5mm}
		\caption{}
		\label{fig:clustered_pca}
	\end{subfigure}
	\\
	\begin{subfigure}[b]{0.49\linewidth}
		\centering
		\include{tikz/initialization_distance}
		\vspace*{-5mm}
		\caption{}
		\label{fig:abstracted_pca}
	\end{subfigure}%
	\begin{subfigure}[b]{0.49\linewidth}
		\centering
 		\includegraphics[width=\linewidth]{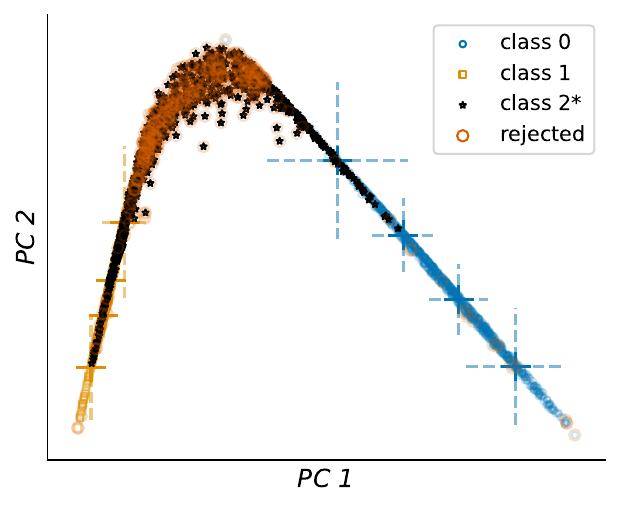}
		\caption{}
		\label{fig:mnist_pca}
	\end{subfigure}
	\caption{Illustration of the steps for initializing the quantitative monitor on a fixed class, in a two-dimensional projection on the first two principal components $PC~1$ and $PC~2$ of the feature layer $\ell$.
	\subref{fig:raw_pca}~Sampling of data points.
	\subref{fig:clustered_pca}~Result of clustering (here: two clusters \cI and \cII) where {\color{niceblue}$\boldsymbol\times$} and {\color{niceorange}$\boldsymbol\times$} respectively mark the cluster centers.
	\subref{fig:abstracted_pca}~Quantitative metric for each cluster, visualized as dashed lines.
	\subref{fig:mnist_pca}~Projection of an initialized quantitative monitor and its detection results for a network trained on the first two classes of the MNIST dataset.}
	\label{fig:construct_monitor}
\end{figure}

In addition to a general framework, we also propose a quantitative monitor for neural networks that fits well into the framework.
In a nutshell, our quantitative monitor works as follows.
At run-time, given an input $\vx$ and a corresponding prediction $\class$ of the neural network, the monitor observes the feature layer $\ell$ and compares its valuation to a model of ``typical'' behavior for the class $\class$.
Next we describe the steps to initialize this monitor, i.e., to construct said behavioral model, which are also illustrated in \figref{construct_monitor}.
Given a labeled training dataset, we observe the neuron valuations \Vals[\class] for each class $\class \in \classes$ (\figref{construct_monitor}\subref{fig:raw_pca}).
We then apply a clustering algorithm to the sets \Vals[\class] (\figref{construct_monitor}\subref{fig:clustered_pca}).
In our implementation, we use a \emph{$k$-means} \cite{Lloyd82} algorithm that finds a suitable $k$ dynamically.

So far the initialization is shared with the qualitative monitor from \cite{henzinger2020outside}, which would next compute the \emph{box abstraction} for each cluster.
A qualitative abstraction-based monitor can only determine whether a point lies inside the abstraction (here: a box) or not.
Since we are interested in a quantitative monitor, we instead define a \emph{distance function} below.

\paragraph{Distance function.}
We set reference points for computing the distance function at the cluster centers.
This way the majority of points have low distance.
Below we describe the particular distance function that we found effective in our evaluation, which we also depict in \figref{construct_monitor}\subref{fig:abstracted_pca} and \subref{fig:mnist_pca}.
We note, however, that the idea generalizes to arbitrary metrics.

Let us fix a class $\class \in \classes$ and a corresponding cluster $\onebox^{\class}$ with center $\vc = (c_1, \dots, c_n)\transpose$ of dimension $n$.
Let $\vr = (r_1, \dots, r_n)\transpose$ be the radius of the bounding box around the cluster.
We define the distance of a point $\vp = (p_1, \dots, p_n)\transpose$ to $\onebox^{\class}$ as the maximum absolute difference to $\vc$ in any projected dimension $i$, normalized by the radius $r_i$:
\begin{equation*}\label{eq:box_distance_single}
    \dbox(\vp, \onebox^{\class}) = \max_{i} |c_i - p_i| \cdot r_i^{-1}.
\end{equation*}

The distance generalizes to a set $\boxes^{\class}$ of clusters for the same class \class by taking the minimum distance in the set:
\begin{equation*}\label{eq:box_distance_set}
    \dbox(\vp, \class) = \min_{\onebox^{\class} \in \boxes^{\class}} \dbox(\vp, \onebox^{\class}).
\end{equation*}

Computing the distance is linear in the dimension (i.e., the number of neurons in the feature layer).
We note that we can in principle also generalize the distance to a set of classes $\classes$ in order to obtain a new classifier.
In this paper, for the purpose of monitoring, we just compare the distance for a fixed class to some class-specific threshold (which we explain later).

\subsection{Active monitoring algorithm}
\label{subsec:algorithm}

\begin{figure}[t]
	\centering
	\includegraphics[width=\textwidth,keepaspectratio]{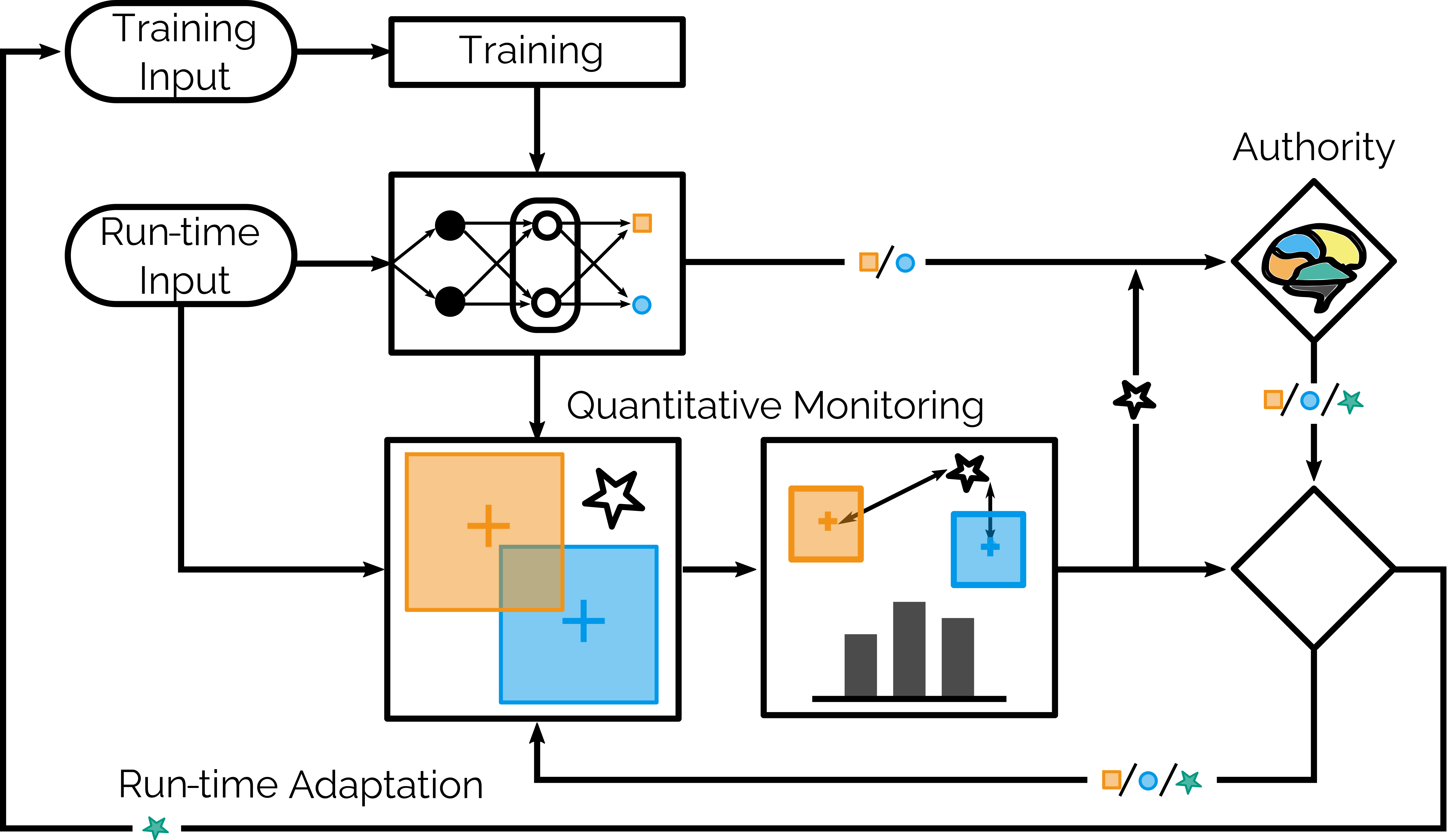}
	\caption{High-level overview of the framework.
	The neural-network classifier receives an input at run-time (top left).
	This input is classified while the monitor watches the classification process.
	The authority is only queried for the correct label if the monitor reports a misclassification.
	That may trigger, depending on the result and the history, adaptation of the monitor or of the model.}
	\label{fig:framework}
\end{figure}

We now explain our active monitoring algorithm, summarized in \algref{framework} and also illustrated in \figref{framework}.

\paragraph{Initialization.} We start with a trained neural network \NN with a feature layer~$\ell$ and a dataset \traindata with a number of classes (the ``known'' classes) as inputs.
The first step in line~\ref{line:start} is to initialize a monitor \M for this network, for example as described in Section~\ref{subsec:monitor_init} for our quantitative monitor.
Instead of working on the feature layer's neurons directly, we learn a transformation matrix by applying principal component analysis (PCA) \cite{Jolliffe86} or Kernel PCA \cite{ScholkopfSM97} to the neuron valuations \Vals[\class].
This transformation is not a requirement of our framework, and hence we omit it in the pseudocode; as we noticed experimentally, this step tends to further separate the valuations \Vals[\class] and \Vals[\class'] for different classes $\class' \neq \class$, which improves the overall monitor precision.

\begin{algorithm}[t]
    \caption{Active monitoring}
    \label{algo:framework}
    \KwIn{%
    \NN: trained model \\
    \traindata: training data \\
    $\traindata_\mathit{run}$: online input stream \\
    }
    \vspace*{2mm}
    \While{True}{
        $\M, \classes \asgn$ \build(\NN, \traindata, $\ell$) \tcp*[h]{build monitor \M and extract known classes \classes from \traindata} \label{line:start} \\
        \While{True}{
            \tcp*[h]{monitoring mode} \\
            $\vx \asgn \get(\traindata_\mathit{run})$ \tcp*[h]{get next input \vx} \label{line:monitoring_start} \\
            $\class \asgn$ \classify(\NN, \vx)\label{line:classify} \tcp*[h]{predict class of \vx} \\
            $\vp \asgn$ \observe(\NN, \vx, $\ell$)\label{line:watch} \tcp*[h]{observe output at layer $\ell$} \\
            $\textit{warning}, \vs \asgn$ \monitor(\vp, $\class$, \M) \tcp*[h]{monitor and compute statistics $\vs$} \\
            \If {\textit{warning}}{
                $\class^{\star} \asgn \askAuthority(\vx,\class,\dbox(\vp, \class))$ \\
                \traindata $\asgn$ \collect($\vx, y^{\star}, \traindata$) \tcp*[h]{add labeled pair $(\vx, y^{\star})$ to \traindata}\\
                \textit{adapt\_model} $\asgn$ \evaluate(\vs, \traindata, \classes) \label{line:monitoring_end} \\
                \tcp*[h]{adaptation mode} \\
                \uIf{\textit{adapt\_model} \label{line:adaptation_start}}{
                    \NN, \M, \traindata $\asgn \adaptModel(\NN,\traindata)$~\rboxed{B}\\
                    \textbf{break}\label{line:adaptation_middle1}
                }
                 \M\,$\asgn$\,\adaptMonitor(\vs,\,\M,\,\NN,\,\traindata)~\rboxed{A}\hspace*{-4mm}\label{line:adaptation_end}
            }
        }
    }
\end{algorithm}

\paragraph{Monitoring stage (lines~\ref{line:monitoring_start}--\ref{line:monitoring_end}).}
At run-time, we apply our framework to a stream of inputs.
For each input \vx, we perform the following steps.
We first apply the neural network to obtain both the class prediction $\class$ and the (principal components of the) neuron valuations \vp at the feature layer~$\ell$.
We then query the monitor \M about the prediction.
In the case of the quantitative monitor, \M computes the distance $\dbox(\vp, \class)$ with respect to the predicted class $y$.
Then \M compares this distance to a class-specific threshold $d^*_{\class}$; initially this threshold is set to $1$, but we increase this value during the course of the algorithm later.

In the simple case that $\dbox(\vp, \class) \leq d^*_{\class}$, the monitor does not raise a warning and the framework just returns the predicted class $\class$ for input \vx (not shown in the pseudocode).
Otherwise the monitor rejects the network prediction as unknown.
In this case we query the authority to provide the ground truth $\class^*$ for input \vx and add the pair $(\vx, \class^*)$ to our training dataset~\traindata.
For our quantitative monitor we additional provide the authority with the distance $\dbox(\vp, \class)$ as a confidence measure, while for qualitative monitors this argument is missing.
The procedure $\evaluate(\vs,\traindata,\classes)$, where $\vs=\{\statnet,\statsample,\statmon\}$, decides between the following two scenarios, which we describe afterward.
\begin{itemize}
	\item[\rboxed{A}]  The ground truth matches the prediction ($\class^* = \class$).
	In this case it was not correct to raise a warning and we continue with the monitor adaptation.

	\item[\rboxed{B}] The ground truth does not match the prediction ($\class^* \neq \class$), possibly because $\class^*$ is unknown to \NN.
	In this case it was correct to raise a warning and we continue with the model adaptation.
 \end{itemize}

\paragraph{Monitor adaptation (line~\ref{line:adaptation_end}).}
Procedure \adaptMonitor(\vs,\,\M,\,\NN,\,\traindata) for monitor adaptation in \rboxed{A} is triggered if a wrong warning was raised and only applies to our quantitative monitor.
Recall that the reason for raising a warning is that the distance of \vp exceeds the threshold for class \class.
We do not immediately adapt the monitor every time it raises a wrong warning.
Instead we keep track of the monitor's performance over time in terms of a score \statmon as defined in Section~\ref{sec:background}.
We only adapt the monitor if \statmon drops below a user-defined threshold \statmonStar.
The adaptation performs two simple steps.
First, we adapt the cluster centers to the new collected data in \traindata.
Second, we adapt the distance threshold $d^*_{\class}$ as follows.
Let \statsample be the number of samples of class $\class$ that we have already collected in \traindata, and let $\statsampleStar$ be a learning threshold as defined in Section~\ref{sec:background}.
We define the new threshold $d^*_{\class}$ (which increases compared to the old value) as
\begin{equation*}
 	d^*_{\class} + (\dbox(\vp, \class) - d^*_{\class}) \cdot \frac{\statsampleStar}{\statsample}.
\end{equation*}

\paragraph{Model adaptation (lines~\ref{line:adaptation_start}--\ref{line:adaptation_middle1}).}
In contrast to monitor adaptation, \emph{model} adaptation in \rboxed{B} involves retraining the neural-network model in order to learn novel classes of inputs.
Procedure $\adaptModel(\NN,\,\traindata)$ performs this adaptation only if one of the following conditions is satisfied:

\begin{itemize}
    \setlength{\itemindent}{.5em}
    \item[\rboxed{B.1}] The number of collected samples labeled by the authority reaches a pre-defined threshold \statsampleStar (see Section~\ref{sec:background}).
    \item[\rboxed{B.2}] The precision score of the current model \statnet falls below the desired value \statnetStar (see Section~\ref{sec:background}).
\end{itemize}

In \rboxed{B.1}, using the dataset \traindata replenished with the data points reported by the monitor and labeled by the authority, we identify which class (or multiple classes) should be learned, based on the collected statistics $\vs.$ 
We then employ transfer learning \cite{PanY10} to train a new model that recognizes this class (classes) in addition to the ones already known.
Specifically, we remove the output layer and all trailing layers until the last fully connected one and then add a new output layer corresponding to the desired number of classes present in \traindata. From the newly compiled model we also augment the monitor.
In the case of our quantitative monitor, we apply the steps from Section~\ref{subsec:monitor_init} for the new class(es) and set the corresponding distance threshold(s) to~$1$.

In \rboxed{B.2}, we rely on regular run-time measurements of the precision score for the current model. Algorithmically, this is achieved by keeping a separate (not used for retraining) test dataset after each successful transfer learning. 
We collect only the inputs reported by our monitor and subsequently labeled by the authority.
This is in line with our main objective for the human in the loop to remain the ultimate trustee for the framework.

\begin{remark}
\label{remark:model}
The model obtained from transfer learning on the accumulated labeled samples is not meant as a replacement for the original model provided at the initialization stage but rather as an assistant to ongoing active monitoring.
\end{remark}

This concludes all possible cases for one iteration of the algorithm. This process is repeated for each input in the stream.

\section{Experiments}
\label{sec:experiments}

We perform two experiments.
In the first experiment we compare our quantitative monitor to three other (static) monitoring strategies: a box-abstraction monitor \cite{henzinger2020outside}, a monitor based on the softmax score \cite{HendrycksG17}, and a monitor that warns with uniform random rate. We evaluate these monitors on five image-classification datasets.
In the second experiment we investigate the influence of different parameters on our quantitative monitor specifically.

\begin{table}[t]
    \caption{\textbf{Dataset and model description.}
	The columns show the number of samples for training and testing, the number of classes in total and initially known to the network, the ID of the network architecture, and the full dimension (i.e., number of neurons) of the monitored layer.}
	\label{tbl:benchmark_description}
	\centering
	\begin{tabular}{@{\hspace*{3mm}} l @{\hspace*{4mm}} r @{\hspace*{4mm}} c @{\hspace*{4mm}} c @{\hspace*{4mm}} c @{\hspace*{2mm}}}
		\toprule
		\multirow{2}{*}{Dataset} & \cc{Dataset size} & \cc{Classes} & Net & Dimension \\
		& \cc{train / test} & \cc{all / init} & ID & full\\
		\midrule
		MNIST & $60{,}000$\ / $10{,}000$ & $10$\ / \phantom{0}$5$ & 1 & \phantom{0}$40$ \\
		FMNIST & $60{,}000$\ / $10{,}000$ & $10$\ / \phantom{0}$5$ & 1 & \phantom{0}$40$ \\
		CIFAR10 & $50{,}000$\ / $10{,}000$ & $10$\ / \phantom{0}$5$ & 3 & $256$ \\
		GTSRB & $39{,}209$\ / $12{,}630$ & $43$\ / $22$ & 2 & \phantom{0}$84$ \\
		EMNIST & $112{,}800$\ / $18{,}800$ & $47$\ / $24$ & 1 & \phantom{0}$40$ \\
		\bottomrule
	\end{tabular}
\end{table}

\subsection{Benchmark Datasets}
We consider the following publicly available datasets, summarized in \tabref{benchmark_description}: MNIST \cite{lecun1998gradient}, Fashion MNIST (FMNIST) \cite{fashionMNIST}, and Extended MNIST (EMNIST) \cite{CohenATS17} consist of 28x28 grayscale images; CIFAR10 \cite{krizhevsky2009learning} and the German Traffic Sign Recognition Benchmark (GTSRB) \cite{StallkampSSI11} consist of 32x32 color images.

For each of these benchmarks we trained two neural-network models: one model trained on all classes, which we refer to as the ``static full'' model, and one model trained on half of the classes, which we refer to as the ``static half'' model. We used VGG16 \cite{ZhangZHS16} pretrained on ImageNet for CIFAR10 and the architectures from \cite{ChengNY19} for MNIST (which we also use for FMNIST and EMNIST) and GTSRB.

\subsection{Experimental Setup}
We let the framework process inputs in batches of size~128. For each dataset we ran our active monitoring framework on reshuffled data five times.

We evaluate our active monitoring framework with four different monitoring strategies, each of which uses the same overall processing within the framework, e.g., the same sequence of samples in the input stream and the same policy for model adaptation.
The strategy based on the softmax score rejects inputs when the score falls below $0.9$.
The random strategy rejects inputs with probability $p = 5$\% (resp.\ $p = 10$\% in the EMNIST experiment).
To make the comparison fair, we limit the number of available authority queries for each strategy to a budget of $p$ (the random rejection probability) percent of the full dataset.
For most of the benchmarks we used PCA and $\statsampleStar$ as explained in Section~\ref{sec:background}.
For CIFAR10, we used Kernel PCA and $\statsampleStar = 0.01\cdot|\traindata|/|\classes|$ instead.

We implemented our framework in Python~3.6 with Tensorflow~2.2 and scikit-learn. We ran all experiments on an i7-8550U@1.80GHz CPU with 32 GB RAM.
The source code and scripts that we used are available online\footnote{\url{https://github.com/VeriXAI/Into-the-Unknown}}.

\subsection{Experimental Results}
\label{subsec:results}

\paragraph{General performance.}

\begin{table}[t]
    \caption{\textbf{Monitor comparison.}
	We compare four different monitoring strategies: quantitative (this paper), box abstraction, softmax score, and random warning.
	For each benchmark we report the interaction limit with the authority, the highest number of learned classes, and the average monitoring precision of five runs.
	The best results per benchmark are marked in \textbf{bold}.}
	\label{tbl:monitor_comparison}
	\centering
	\begin{filecontents*}{data/experiment2.csv}
Dataset;Limit;QuantitativeLearned;QuantitativeTest;QuantitativeTestPM;AbstractionLearned;AbstractionTest;AbstractionTestPM;SoftmaxLearned;SoftmaxTest;SoftmaxTestPM;RandomLearned;RandomTest;RandomTestPM
MNIST;\phantom{0}3{,}000;\textbf{10};\textbf{0.81};0.01;\textbf{10};0.6;0.02;\textbf{10};0.71;0.01;\phantom{0}6;0.48;0.01
FMNIST;\phantom{0}3{,}000;\phantom{0}9;\textbf{0.74};0.02;\phantom{0}9;0.54;0.02;\textbf{10};0.7;0.01;\phantom{0}8;0.5;0.01
CIFAR10;\phantom{0}2{,}500;\textbf{10};\textbf{0.75};0.02;\textbf{10};0.61;0.02;\textbf{10};0.53;0.01;\textbf{10};0.41;0.01
GTSRB;\phantom{0}1{,}960;37;0.67;0.02;\textbf{38};0.7;0.01;34;\textbf{0.75};0.03;25;0.29;0.01
EMNIST;11{,}280;42;\textbf{0.81};0.01;\textbf{47};0.71;0.02;\textbf{47};0.69;0.01;\textbf{47};0.39;0.01
\end{filecontents*}
\csvreader[separator=semicolon,
		tabular=@{\hspace*{3mm}} l @{\hspace*{4mm}} c @{\hspace*{4mm}} r @{\,} c @{\,} l @{\hspace*{4mm}} r @{\,} c @{\,} l @{\hspace*{4mm}} r @{\,} c @{\,} l @{\hspace*{4mm}} r @{\,} c @{\,} l,
		table head=\toprule
		\multirow{2}{*}{Dataset} & Interaction & \multicolumn{3}{@{} c @{\hspace*{4mm}}}{Quantitative\hspace*{-1mm}} & \multicolumn{3}{@{} c @{\hspace*{4mm}}}{Abstraction\hspace*{-0.5mm}} & \multicolumn{3}{@{} c @{\hspace*{4mm}}}{Softmax} & \multicolumn{3}{@{} c @{\hspace*{4mm}}}{Random} \\
		& limit & \multicolumn{3}{@{} c @{\hspace*{4mm}}}{class/prec} & \multicolumn{3}{@{} c @{\hspace*{4mm}}}{class/prec} & \multicolumn{3}{@{} c @{\hspace*{4mm}}}{class/prec} & \multicolumn{3}{@{} c @{\hspace*{4mm}}}{class/prec} \\
		\midrule,
		table foot=\bottomrule]
		{data/experiment2.csv}
		{1=\Kdataset,2=\Klimit,3=\Kql,4=\Kqt,5=\Kqtpm,6=\Kal,7=\Kat,8=\Katpm,9=\Ksl,10=\Kst,11=\Kstpm,12=\Krl,13=\Krt,14=\Krtpm}
		{\multirow{2}{*}{\Kdataset} & \multirow{2}{*}{\Klimit} & \multicolumn{3}{c}{\Kql} & \multicolumn{3}{c}{\Kal} & \multicolumn{3}{c}{\Ksl} & \multicolumn{3}{c}{\Krl} \\
		&& \Kqt & $\pm$ & \Kqtpm & \Kat & $\pm$ & \Katpm & \Kst & $\pm$ & \Kstpm & \Krt & $\pm$ & \Krtpm}
\end{table}

\begin{figure}[p]
	\centering
	\begin{subfigure}[b]{.49\linewidth}
		\centering
		\includegraphics[width=\linewidth]{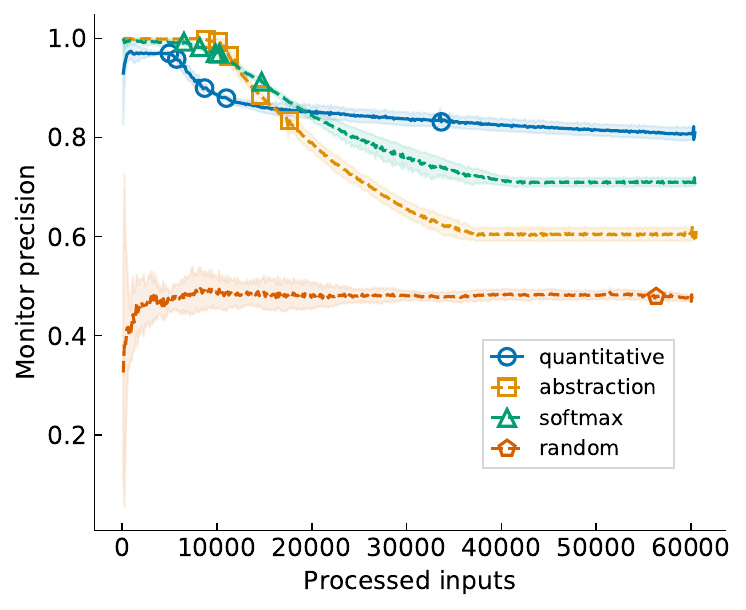}
		\caption{MNIST}
	\end{subfigure}%
	\hfill
	\begin{subfigure}[b]{.49\linewidth}
		\centering
		\includegraphics[width=\linewidth]{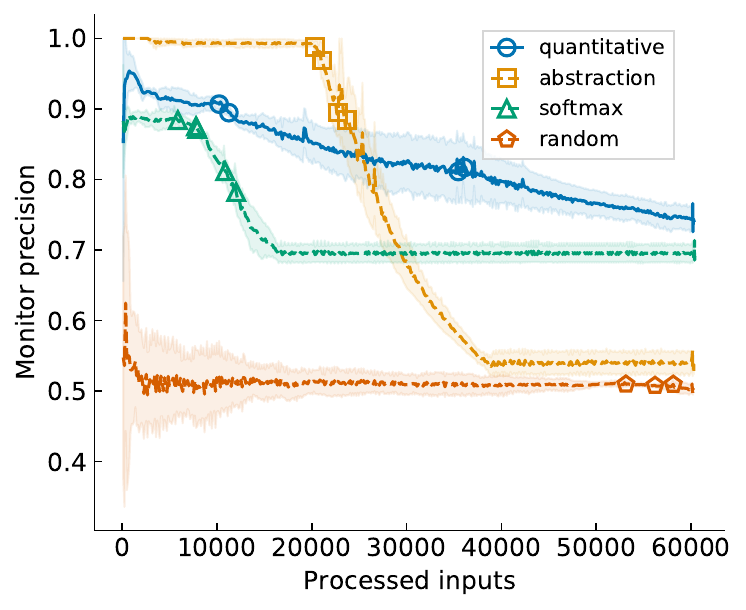}
		\caption{FMNIST}
	\end{subfigure}%
	
	\begin{subfigure}[b]{.49\linewidth}
		\centering
		\includegraphics[width=\linewidth]{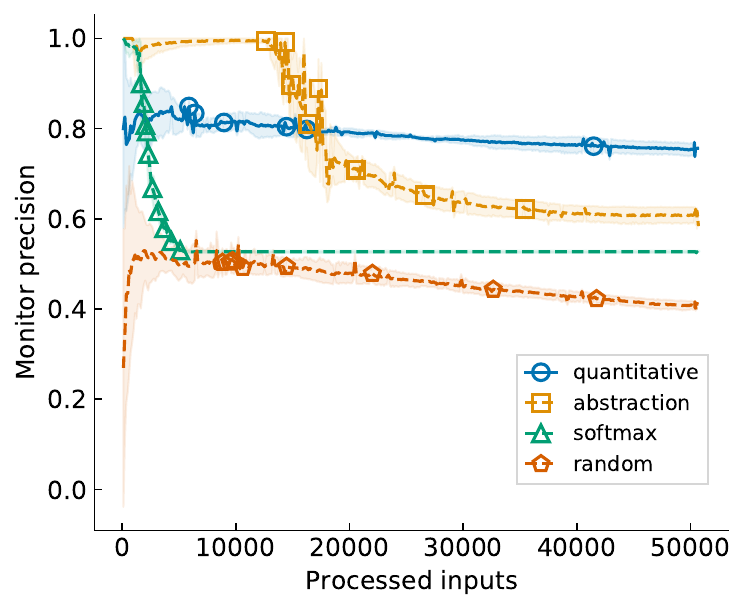}
		\caption{CIFAR10}
	\end{subfigure}%
	\hfill
	\begin{subfigure}[b]{.49\linewidth}
		\centering
		\includegraphics[width=\linewidth]{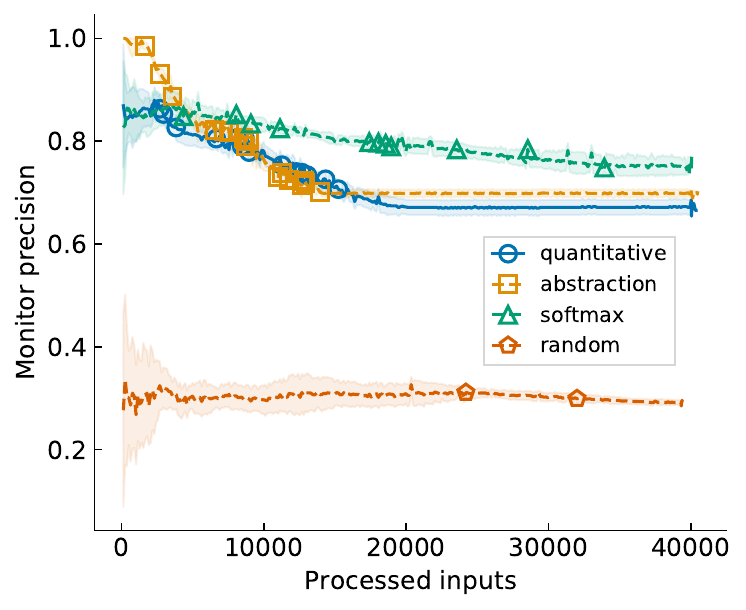}
		\caption{GTSRB}
	\end{subfigure}%
	
	\begin{subfigure}[b]{.49\linewidth}
		\centering
		\includegraphics[width=\linewidth]{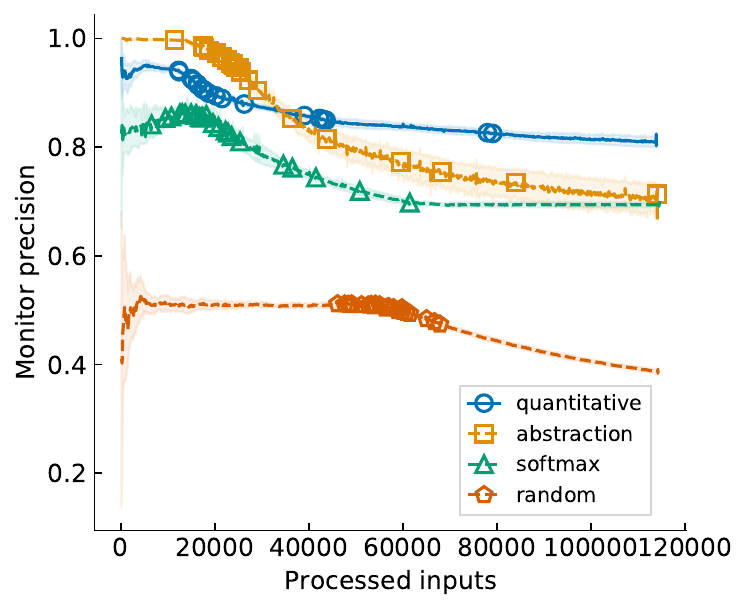}
		\caption{EMNIST}
	\end{subfigure}%
	\caption{Comparison of the monitor precision between four monitoring strategies, averaged over five runs and including 95\%-confidence bands. The markers correspond to points in time when a model adaptation takes place.}
	\label{fig:precision_monitor}
\end{figure}

The performance of the different monitoring strategies in terms of monitoring precision is averaged over five runs and summarized in \tabref{monitor_comparison}. For all but one benchmark our monitor achieves the highest precision, and for GTSRB the precision is comparable with other monitors. \figref{precision_monitor} shows the evolution of the monitor precision over time as more classes are learned.
Recall that the network is dynamically retrained (using transfer learning) for new classes.
Clearly, the number of new samples for this training procedure is lower than in a normal, full-fledged training.
Consequently, the adapted network is less precise for these new classes (cf.\ \tabref{active_learning}) than a network trained on the full training dataset.
Hence it is expected that the general trend in the monitoring precision is decreasing for all strategies.

\begin{table}[t]
	\caption{\textbf{Model adaptation.}
	We compare the static model trained on 50\% of the classes, the static model trained on all classes, and the model obtained from our framework (using the quantitative monitor), averaged over five runs. In the static cases, the test accuracy is measured on the filtered test set (not including novelties for the 50\% model).
	The second column shows the epochs used for the initial training resp.\ the retraining/transfer learning at run-time.
	}
	\label{tbl:active_learning}
	\centering
	\begin{filecontents*}{data/experiment1.csv}
Dataset;EpochsTrain;EpochsRun;FullAccTraining;FullAccValidation;HalfAccTraining;HalfAccValidation;AdaptiveTest;AdaptiveTestPM
MNIST;10;10;0.99;0.99;0.99;0.99;0.97;0.01
FMNIST;10;10;0.97;0.91;0.99;0.92;0.79;0.05
CIFAR10;50;30;0.99;0.79;0.99;0.83;0.54;0.02
GTSRB;30;30;0.99;0.88;0.99;0.95;0.87;0.01
EMNIST;30;30;0.92;0.86;0.97;0.92;0.71;0.04
\end{filecontents*}
\csvreader[separator=semicolon,
		tabular=@{\hspace*{3mm}} l @{\hspace*{4mm}} c @{\hspace*{4mm}} c @{\hspace*{4mm}} c @{\hspace*{4mm}} r c l @{\hspace*{3mm}},
		table head=\toprule
		\multirow{2}{*}{Dataset} & Epochs & Static half & Static full & \multicolumn{3}{@{\hspace*{-2.5mm}} c}{Adaptive} \\
		& init/run & train / test & train / test & \multicolumn{3}{@{\hspace*{-2.5mm}} c}{test} \\ \midrule,
		table foot=\bottomrule]
		{data/experiment1.csv}
		{1=\Kdataset,2=\Kepochstrain,3=\Kepochsrun,4=\Kofflinetrain,5=\Kofflinetest,6=\Kstatictrain,7=\Kstatictest,8=\Kadaptivetest,9=\Kadaptivetestpm}
		{\Kdataset & \Kepochstrain \ / \Kepochsrun & \Kstatictrain \ / \Kstatictest & \Kofflinetrain \ / \Kofflinetest & \Kadaptivetest & $\pm$ & \Kadaptivetestpm}
\end{table}

We report the test accuracy of the neural networks in \tabref{active_learning}, averaged over five runs per benchmark. The accuracy is generally lower than what could be achieved by training the network with a full and balanced dataset from scratch (the ``static full'' model), but for some benchmarks we achieve almost the same accuracy. This shows that the framework is able to adapt to new situations.

\paragraph{Cost analysis.}

\begin{figure}[p]
	\centering
	\begin{subfigure}[b]{.49\linewidth}
		\centering
		\includegraphics[width=\linewidth]{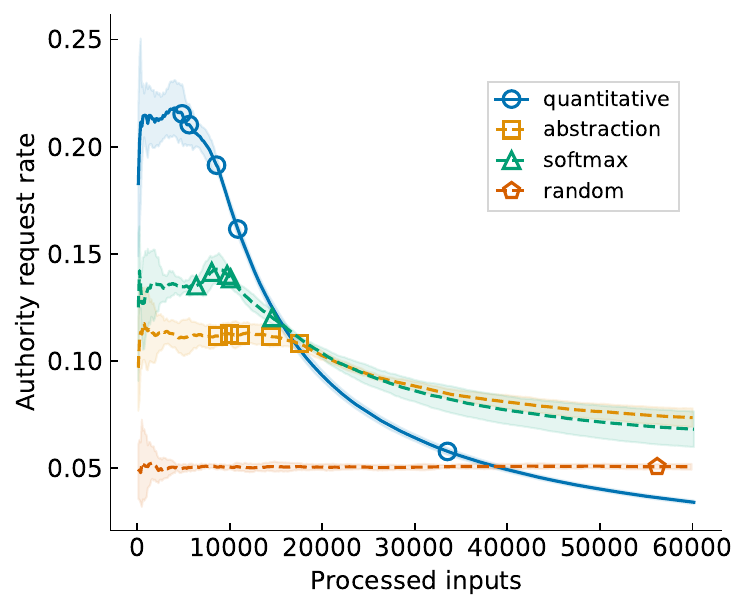}
		\caption{MNIST}
	\end{subfigure}%
	\hfill
	\begin{subfigure}[b]{.49\linewidth}
		\centering
		\includegraphics[width=\linewidth]{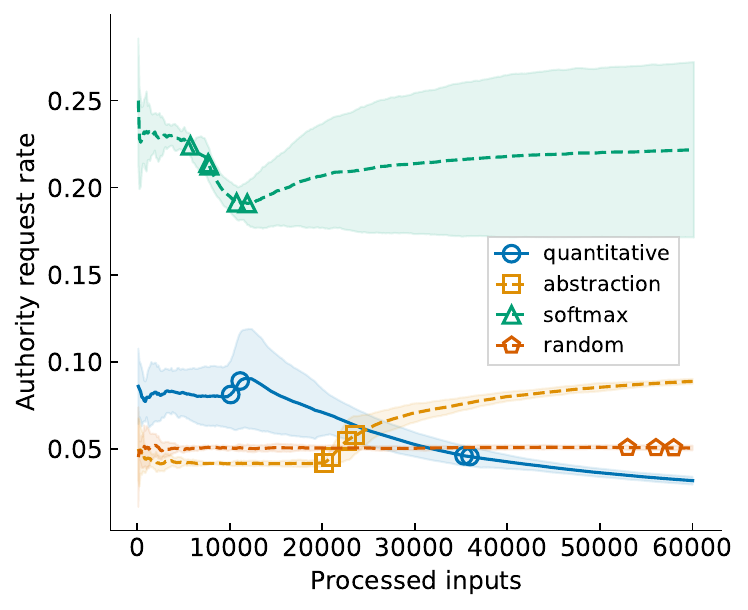}
		\caption{FMNIST}
	\end{subfigure}%
	
	\begin{subfigure}[b]{.49\linewidth}
		\centering
		\includegraphics[width=\linewidth]{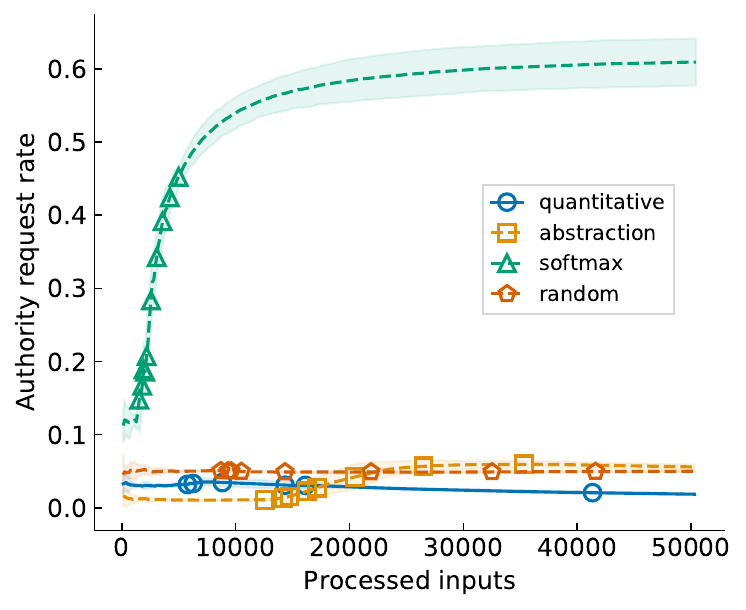}
		\caption{CIFAR10}
	\end{subfigure}%
	\hfill
	\begin{subfigure}[b]{.49\linewidth}
		\centering
		\includegraphics[width=\linewidth]{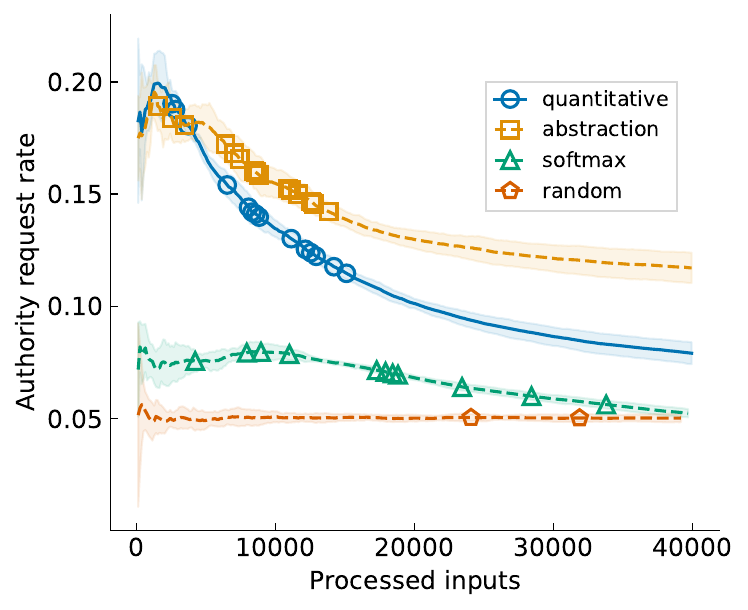}
		\caption{GTSRB}
	\end{subfigure}%
	
	\begin{subfigure}[b]{.49\linewidth}
		\centering
		\includegraphics[width=\linewidth]{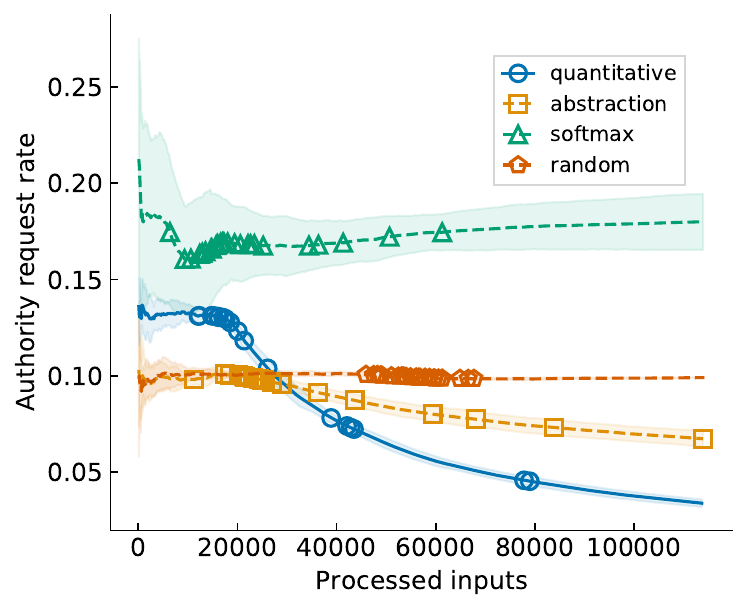}
		\caption{EMNIST}
	\end{subfigure}%
	\caption{Comparison of the rate of authority queries between four monitoring strategies, averaged over five runs and including 95\%-confidence bands. The markers correspond to points in time when a model adaptation takes place.}
	\label{fig:requests}
\end{figure}

In \figref{requests}, we show the frequency of authority queries over time. Recall that there is a budget of queries (cf.\ \tabref{monitor_comparison}). Our quantitative monitor queries the authority more frequently at the beginning but as it adapts to more novel classes the rate of requests is steadily decreasing. Thus the monitor has the fewest queries in four of the five benchmarks (except for GTSRB). The other monitors do not have an adaptation mechanism and therefore are prone to querying the authority more often. For some monitors we even observe an increase in warnings over time, in particular the monitor that uses the softmax score. As we argued above, we suspect that the network tends to be less confident for newly learned classes, which results in lower softmax scores.
Learning new classes often happens at roughly the same point in time. This is because the novelties appear with uniform distribution in the input stream; hence the points in time when a fixed number per class has been seen are close to each other.

Overall the plots do not reveal a clear trend which monitor is fastest at learning new classes. There is generally a trade-off between the rate at which a warning is raised and the rate at which new classes are learned. In our scenario, raising a warning is initially correct in 50\% of the cases (note that none of the monitors is in that range); taken to the extreme, a monitor that always raises a warning would be the fastest in learning new classes. On the other hand, a monitor that generally raises fewer warnings to the authority may also miss novelties and thus learn slower. However, in our experience it is more preferred to provide a low false-positive rate, i.e., warnings raised by the monitor should be genuine. In this sense the quantitative monitor works best.

\begin{figure}[t]
	\centering
	\begin{subfigure}[b]{.49\linewidth}
		\centering
		\includegraphics[width=\linewidth]{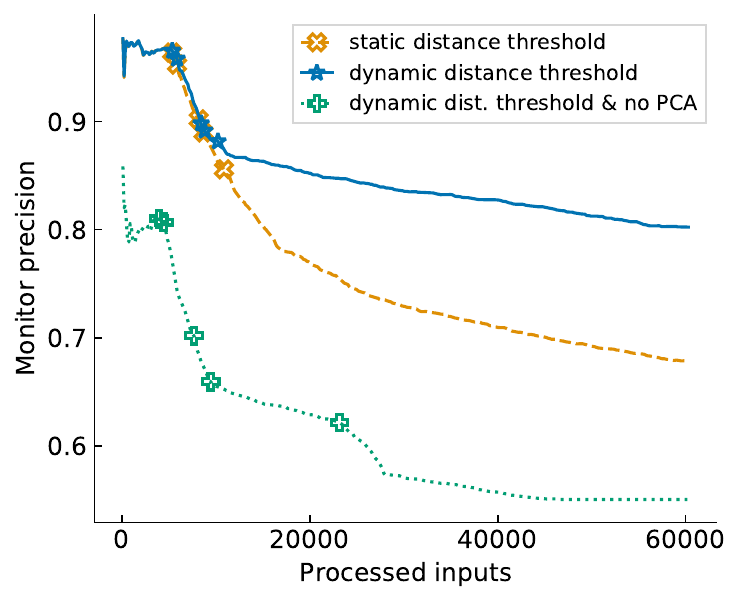}
		\caption{Static and dynamic distance threshold.}
		\label{fig:distance_one}
	\end{subfigure}%
	\hfill
	\begin{subfigure}[b]{.49\linewidth}
		\centering
		\includegraphics[width=\linewidth]{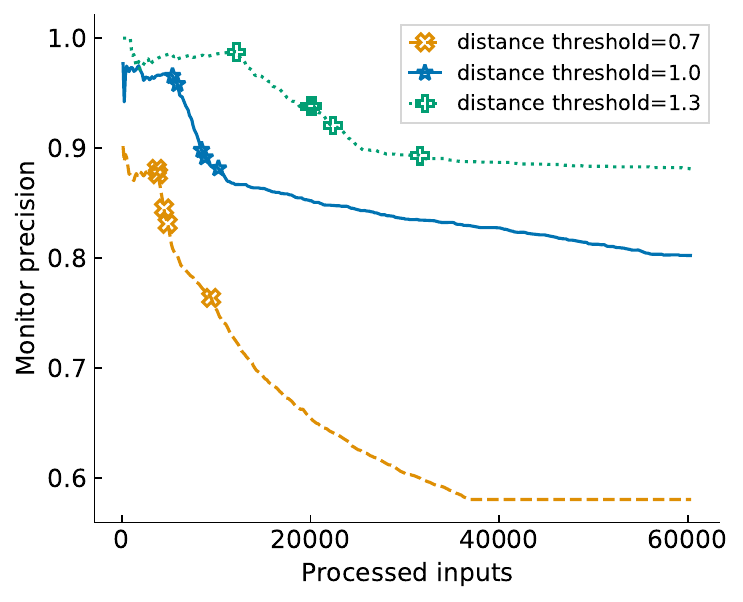}
		\caption{Different initial threshold values.}
		\label{fig:distance}
	\end{subfigure}%
	\caption{Influence of the dynamic distance threshold $d^*(\class)$ for each class \class on the quantitative-monitor precision for the MNIST benchmark.
	The markers correspond to points in time when a model adaptation takes place.
	\subref{fig:distance_one}~Comparison between a static value and a dynamically changing value (as proposed in this paper); we also show a comparison with a run where we omit the preprocessing with PCA.
	\subref{fig:distance}~Influence of the initial value of the threshold.}
	\label{fig:distance_comparison}
\end{figure}

\paragraph{Ablation and sensitivity study.}
All components of our framework contributed to its performance. In \figref{precision_monitor}, we have illustrated how incremental retraining of the model improves the monitor precision for all monitoring strategies. In principle, other active-learning strategies can be plugged into our framework to further increase this effect. In addition, \figref{precision_monitor} demonstrates that the monitor-adaptation stage (where the monitor is incrementally adjusted without \emph{model} adaptation), which only applies to our quantitative monitor, helps maintaining a better precision than the other monitoring strategies.

\figref{distance_comparison}\subref{fig:distance_one} shows that dynamically changing the value of the distance threshold $d^*(\class)$ (for each class \class) contributes to the precision of our monitor, and so does the use of PCA for dimensionality reduction.
Similarly, \figref{distance_comparison}\subref{fig:distance} shows that the starting value of the (dynamic) threshold also influences the monitor precision.

\begin{table}[t]
	\caption{\textbf{Average run times in seconds.}
	For each benchmark we average (five runs) the time for retraining the neural network (when enough samples of a new class were collected),
	for retraining the monitor (after retraining the neural network),
	and for adapting the monitor (when the precision drops too much).}
	\label{tbl:runtimes}
	\centering
	\begin{filecontents*}{data/runtimes.csv}
Dataset;RetrainNet;RetainNetPM;RetrainMon;RetrainMonPM;Adapt;AdaptPM
MNIST;\phantom{0}26;\phantom{00}1;\phantom{0,0}59;\phantom{00}3;39;\phantom{0}5
FMNIST;\phantom{0}19;\phantom{00}6;\phantom{0,0}45;\phantom{0}10;59;\phantom{0}5
CIFAR10;257;\phantom{0}57;2{,}477;282;40;\phantom{0}3
GTSRB;228;\phantom{0}12;194;\phantom{0}24;19;\phantom{0}1
EMNIST;360;191;347;\phantom{0}71;82;16
\end{filecontents*}
\csvreader[separator=semicolon,
		tabular=@{\hspace*{3mm}} l @{\hspace*{4mm}} r @{\,} c @{\,} l @{\hspace*{4mm}} r @{\,} c @{\,} l @{\hspace*{4mm}} r @{\,} c @{\,} l @{\hspace*{3mm}},
		table head=\toprule
		\multirow{2}{*}{Dataset} & \multicolumn{3}{@{\ } c @{\hspace*{4mm}}}{Retrain} & \multicolumn{3}{@{\ } c @{\hspace*{4mm}}}{Retrain} & \multicolumn{3}{@{\hspace*{-2mm}} c}{Adapt} \\
		& \multicolumn{3}{@{\ } c @{\hspace*{4mm}}}{network} & \multicolumn{3}{@{\ } c @{\hspace*{4mm}}}{monitor} & \multicolumn{3}{@{\hspace*{-2mm}} c}{monitor} \\
		\midrule,
		table foot=\bottomrule]
		{data/runtimes.csv}
		{1=\Kdataset,2=\Kretrainnet,3=\Kretrainnetpm,4=\Kretrainmon,5=\Kretrainmonpm,6=\Kadapt,7=\Kadaptpm}
		{\Kdataset & \Kretrainnet & $\pm$ & \Kretrainnetpm & \Kretrainmon & $\pm$ & \Kretrainmonpm & \Kadapt & $\pm$ & \Kadaptpm}
\end{table}

\paragraph{Timing analysis.}

\tabref{runtimes} shows a timing comparison for the individual adaptation stages of the framework, taken from the runs for the quantitative monitor strategy.
(Comparing different strategies is generally difficult because they interact with the authority and adapt the model and/or the monitor in different orders and frequencies.)
The time grows with the size of the dataset but on average is on the order of milliseconds per input; hence the framework can be run in real time.
For CIFAR10 the time is dominated by the use of Kernel PCA.

\section{Conclusion and future work}
\label{sec:discussion}
In this work, we have presented an active monitoring framework for accompanying a neural-network classifier during deployment. The framework adapts to unknown input classes via interaction with a human authority. Experiments on a diverse set of image-classification benchmarks showed that active monitoring is effective in improving accuracy over time in the setting when inputs of novel classes are frequently encountered. Moreover, we introduced a new quantitative monitor, providing the human with confidence about the reported warnings based on a distance to the predicted class in feature space. In comparison to alternative monitoring strategies, our monitor demonstrated superior performance in detection and adaptation at run-time. Our framework thus improves trustworthiness of automated decision making.

Our framework is independent of the choice of the dataset and the neural-network architecture. The only requirements for applicability of our approach are access to the output of the feature layer(s). We plan to extend our procedure toward real-world applications with particular need of active monitoring, e.g., in robotics for the trained controller to gradually adapt to the behavior of the authority. Other interesting directions are time-critical applications where the adaptation of the monitor or the neural network need to be delayed to uncritical phases, and scenarios where novel inputs occur rarely. In addition, the underlying method of our framework can serve as a suitable tool for designing an algorithmic approach to explainability of a neural network's predictions.

\section*{Acknowledgments}

We thank Christoph Lampert and Alex Greengold for fruitful discussions.
This research was supported in part by the Simons Institute for the Theory of Computing, the Austrian Science Fund (FWF) under grant Z211-N23 (Wittgenstein Award), and the European Union's Horizon 2020 research and innovation programme under the Marie Sk{\l}odowska-Curie grant agreement No.\ 754411.

\bibliographystyle{splncs04}
\bibliography{bibliography}
\end{document}

%% file: tikz/initialization_raw.tex
\newcommand{\axes}{%
	\coordinate (o) at (0,0);
	\coordinate (1x) at (2.5,0);
	\coordinate (1y) at (0,2.2);
	\draw (o) -- ($(1x)+(0.4,0)$) node[below,xshift=-24mm] {\scriptsize $\mathit{PC~1}$};
	\draw (o) -- ($(1y)+(0,0.4)$) node[above,yshift=-21mm,rotate=90] {\scriptsize $\mathit{PC~2}$};
}
\newcommand{\points}[3]{%
	\node[#1] (p1) at (0.7,2) {};
	\node[#1] (p2) at (0.6,1.8) {};
	\node[#1] (p3) at (1,1.9) {};
	\node[#1] (p4) at (0.7,1.6) {};
	\node[#1] (p5) at (1,1.7) {};
	\node[#1] (p6) at (1.5,1.85) {};
	\node[#2] (q1) at (2.5,1) {};
	\node[#2] (q2) at (2.3,.9) {};
	\node[#2] (q3) at (2.1,.8) {};
	\node[#2] (q4) at (2.4,.8) {};
	\node[#2] (q5) at (2.2,.6) {};
	\node[#2] (q6) at (2,.5) {};
	\node[#2] (q7) at (2.5,.6) {};
	\coordinate (cp) at (0.9,1.8);
	\coordinate (cq) at (2.25,0.75);
}
\newcommand{\boxdistance}[4]{%
	\draw[color=#4,dashed,very thick] ($(#1) + (-#2,0)$) -- ($(#1) + (#2,0)$);
	\draw[color=#4,dashed,very thick] ($(#1) + (0,-#3)$) -- ($(#1) + (0,#3)$);
	\draw[color=#4,opacity=0.5,very thick] ($(#1) + (-#2,0)$) -- ($(#1) + (#2,0)$);
	\draw[color=#4,opacity=0.5,very thick] ($(#1) + (0,-#3)$) -- ($(#1) + (0,#3)$);
}

%% file: tikz/initialization_pca.tex
\begin{tikzpicture}[scale=1.7]
	\axes
	\points{p}{p}{1}
\end{tikzpicture}

%% file: tikz/initialization_clustered.tex
\begin{tikzpicture}[scale=1.7]
	\axes
	\points{p1}{p2}{2}
	\node[cross,color=niceorange] at (cp) {};
	\node[cross,color=niceblue] at (cq) {};
\end{tikzpicture}

%% file: tikz/initialization_distance.tex
\begin{tikzpicture}[scale=1.7]
	\axes
	\points{p1,opacity=0.2}{p2,opacity=0.2}{3}
	\node[cross,color=niceorange,opacity=0.1] at (cp) {};
	\node[cross,color=niceblue,opacity=0.1] at (cq) {};
	\boxdistance{cp}{0.45}{0.27}{niceorange}
	\boxdistance{cq}{0.35}{0.35}{niceblue}
\end{tikzpicture}